\pdfoutput=1

\documentclass[11pt]{article}

\usepackage[]{acl}

\usepackage{times}
\usepackage{latexsym}
\usepackage{dblfloatfix}

\usepackage[T1]{fontenc}

\usepackage[utf8]{inputenc}

\usepackage{microtype}

\usepackage{inconsolata}

\usepackage{booktabs}
\usepackage{multirow}
\usepackage{multicol}
\usepackage{graphicx}
\usepackage{enumitem}
\usepackage{longtable}

\title{Split and Rephrase with Large Language Models}

\author{David Ponce$^{*1,2}$ \and Thierry Etchegoyhen$^{*1}$ \and Jesús Calleja$^{1,2}$ \and Harritxu Gete$^{1,2}$ \\
        $^1$ Fundación Vicomtech, Basque Research and Technology Alliance (BRTA) \\
        $^2$ University of the Basque Country UPV/EHU \\
        \texttt{\{adponce,tetchegoyhen,jcalleja,hgete\}@vicomtech.org} }

\begin{document}
\maketitle

\def\thefootnote{*}\footnotetext{Equal contribution.}\def\thefootnote{\arabic{footnote}}


\begin{abstract}
The Split and Rephrase (SPRP) task, which consists in splitting complex sentences into a sequence of shorter grammatical sentences, while preserving the original meaning, can facilitate the processing of complex texts for humans and machines alike. It is also a valuable testbed to evaluate natural language processing models, as it requires modelling complex grammatical aspects. In this work, we evaluate large language models on the task, showing that they can provide large improvements over the state of the art on the main metrics, although still lagging in terms of splitting compliance. Results from two human evaluations further support the conclusions drawn from automated metric results. We provide a comprehensive study that includes prompting variants, domain shift, fine-tuned pretrained language models of varying parameter size and training data volumes, contrasted with both zero-shot and few-shot approaches on instruction-tuned language models. Although the latter were markedly outperformed by fine-tuned models, they may constitute a reasonable off-the-shelf alternative. Our results provide a fine-grained analysis of the potential and limitations of large language models for SPRP, with significant improvements achievable using relatively small amounts of training data and model parameters overall, and remaining limitations for all models on the task.

\end{abstract}

\section{Introduction}


Transforming complex sentences into a sequence of shorter sentences, while preserving the original meaning, can facilitate the processing of complex texts for humans and machines alike. It is, for instance, an important operation in text simplification \cite{zhu2010monolingual,narayan-gardent-2014-hybrid,shardlow2014survey}, where simpler split sentences can provide content that is easier to read and comprehend. This task, commonly referred to as Split and Rephrase (SPRP) \citep{narayan-etal-2017-split}, also provides a relevant framework to evaluate natural language processing models, as complex linguistic properties need to be tackled to provide optimal splits and appropriate rephrasing. 

The SPRP task is typically performed via dedicated transformation models, e.g. sequence-to-sequence modelling with a copy mechanism \citep{aharoni-goldberg-2018-split} or graph-based neural segmentation \citep{gao-etal-2021-abcd}. The main limitation for this type of approach is the dependency on training datasets of paired complex and split sentences, which are scarce across languages and domains. As an alternative, \citet{kim-etal-2021-BiSECT} exploit one-to-many alignments in parallel corpora and pivot machine translation (MT) to create an aligned corpus of complex and split sentences, which can in turn be exploited by sequence-to-sequence models. For this type of approach, the main bottleneck is the dependency on parallel corpora with sufficient many-to-one alignments and on quality pivot machine translation models.

Large language models (LLMs), such as the GPT models \citep{gpt} based on the Transformer architecture \citep{vaswani2017transformer}, have demonstrated their strong potential on a large number of downstream tasks \citep{radford2019language,brown2020language}. 
In this work, we measure their performance on the SPRP task, prompting the models to generate split and rephrase hypotheses from complex sentences. We evaluate different variants of the approach on datasets of varying complexity, contrasting zero shot, few shot in-context learning, and fine-tuning scenarios. We notably measure the impact of parameter and training data size, domain shift, and prompt optimisation.

Overall, our results demonstrate that LLMs outperform the previous state of the art by large margins across multiple metrics, with remaining relative deficiencies, notably in terms of split generation. Results from two human evaluations, a comparative 3-way ranking task and a qualitative evaluation, further support our conclusions. 
Our main contributions can be summarised as follows: 

\begin{itemize}

    \item New state of the art on the SPRP task, established on publicly available datasets of varying complexity, over multiple metrics and two human evaluations. 
    
    \item A comprehensive evaluation of LLMs for SPRP, covering LLM variants with and without instruction tuning, along with the impact of different prompting strategies, parameter size, training data volumes, and domain shift. 

    \item Empirical results on LLM strengths and limitations for SPRP, in particular over splitting compliance and rephrasing variability.  
    
\end{itemize}

\section{Related Work}


The Split and Rephrase task was first proposed by \citet{narayan-etal-2017-split}, who created the WebSplit dataset from RDF semantic tuples and provided benchmarks with statistical MT models and LSTMs with attention. \citet{aharoni-goldberg-2018-split} identified data representation in this dataset and provided enhancements along with a new benchmark augmenting the LSTM with a copy mechanism \citep{gu-etal-2016-incorporating}. To address some limitations of WebSplit, notably in terms of vocabulary and naturality, \citet{botha-etal-2018-learning}  presented WikiSplit, a corpus created using Wikipedia data that takes advantages of edits made on a page to split or fuse sentences. They showed that models trained on WikiSplit improved those trained on WebSplit in terms of BLEU \cite{papineni-etal-2002-bleu}. However, they also noted that WikiSplit is a relatively noisy dataset, not suited for evaluation.

Among other approaches, \citet{das2018novel} leveraged dependency parsing information for the SPRP task, whereas \citet{niklaus-etal-2019-transforming} employed handcrafted rules, recursively applied to complex sentences to generate simpler ones. Recent approaches have exploited neural modelling, in particular the Transformer architecture. Thus, \citet{malmi-etal-2019-encode} proposed a tagging approach using a BERT model \citep{devlin-etal-2019-bert} as encoder and a single Transformer decoder layer. \citet{stahlberg-kumar-2020-seq2edits} also explored a tagging model, at the span level instead of the token level, achieving improvements with Pointer Networks \citep{vinyals2015pointernetwork} for end of span prediction and a two-layer Transformer decoder. \citet{berlanga-neto-ruiz-2021-split} introduced a cross-lingual model based on part-of-speech tags, GRU sequence-to-sequence modelling, and BERT for next word prediction. Recently, \citet{alajlouni2023knowledge} reported improvements via transfer knowledge of data generated by the DISSIM rule-based system \citep{niklaus-etal-2019-transforming} to fine-tune an encoder-decoder T5 model .


In this work, we compare our LLM-based variants with two recent approaches and associated corpora. First, we address BiSECT \citep{kim-etal-2021-BiSECT}, a method and corpus based on extracting 1-2 alignments from parallel bilingual corpora and pivot machine translation. BiSECT achieved higher quality than WikiSplit, and sequence-to-sequence models trained on the corpus improved on the SPRP task. 
Additionally, we compare our work to \citet{gao-etal-2021-abcd}, who proposed the ABCD framework where SPRP is performed via neural graph segmentation, according to the number of predicates in the complex sentence, approaching or outperforming parsing-based and encoder-decoder models. They also provided two new datasets for the SPRP task: MinWiki, a reduced version of WikiSplit where only pairs of complex and simple sentences with a matching number of predicates are kept, and DeSSE, a corpus of students' opinion essays in social science consisting of complex sentences and corresponding splits; we used the latter in our experiments, in addition to BiSECT.

Finally, we used the Wiki-BM corpus described in \citet{zhang-etal-2020-small}, who found that the standard WebSplit dataset contained easily identifiable syntactic cues, which even a simple rule-based approach could exploit to match the performance of state-of-the-art models. Their benchmarks feature higher syntactic diversity to overcome this issue and are thus suited for our qualitative evaluation of LLM-based SPRP hypotheses. 

\section{Methodology}
\label{sec:method}

The overall approach to SPRP with LLM is simple: probe a LLM to generate sequences of meaning-preserving sentences from a single complex sentence. We explored two main variants for the task.

We first fine-tuned pretrained LLMs on pairs of complex and split sentences. The objective in this scenario is twofold: (i) measure the performance of bare LLMs on the SPRP task, and (ii) evaluate LLM accuracy with varying number of parameters. For these models, we also evaluate different approaches to prompting, comparing soft and manually crafted prompts.


Additionally, we evaluate scenarios where we directly prompt LLMs that have been previously tuned on instruction datasets. The goal in this case is to measure the ability of this type of readily-trained models to perform the two tasks at hand, with or without positive examples. Under this scenario, we explore both zero-shot and in-context few-shot task completion, by directly prompting the model for completion in the first case, and providing a limited number of examples within the prompt in the latter case. 


\section{Experimental Setup}

\subsection{Corpora} 
\label{subsec:corpora}

To train and evaluate our models we first selected the aforementioned English datasets: DeSSE and BiSECT.\footnote{We use the datasets made available in the following repositories, as of April 2023: https://github.com/serenayj/DeSSE and https://github.com/mounicam/BiSECT.}  Statistics for each corpus are provided in Table~\ref{table:corpora}, in terms of number of pairs of complex and split sentences. We indicate the figures for the original dataset (\textit{Original})  and the versions we employed in this work (\textit{Selected}).

 We excluded all pairs where the split sentence did not contain at least one inner end-of-sentence punctuation mark (286 instances in DeSSE and 3 in BiSECT). For training, we down-sampled BiSECT to the size of DeSSE for comparison purposes. As development data, since the original sets for DeSSE were rather small (42 examples), we sampled 2,000 pairs from both training datasets.

\begin{table}[]
\centering
\begin{tabular}{llrrrr}
\toprule
Corpus & Version & Train & Dev & Test \\
\midrule
DeSSE & Original & 13,199 & 42 & 790 \\
DeSSE & Selected & 6,432 & 2,000 & 504 \\
\midrule
BiSECT & Original & 928,440 & 9,079 & 583 \\
BiSECT & Selected & 6,432 & 2,000 & 580 \\
\bottomrule
\end{tabular}
\caption{DeSSE and BiSECT English corpora statistics}
\label{table:corpora}
\end{table}



We chose these two corpora as they present different challenges for SPRP. In terms of topics, DeSSE is based on social science opinion essays and is therefore less varied than BiSECT, which was created from a large variety of parallel corpora. In terms of complex sentence average length, the two datasets are also markedly different: for training, the average is 40.14 tokens in BiSECT, contrasting with 25.25 in DeSSE; in the test sets, the averages were 39.55 and 24.10, respectively.

Finally, for our qualitative manual evaluation, we used the Wiki-BM corpus \citet{zhang-etal-2020-small}, as it was manually crafted for the SPRP task.

\subsection{Models}
\label{sec:models}

\paragraph{Pretrained language models.} As pretrained LLMs, we used the Pythia suite \citep{pythia}, a collection of Transformer-based decoder-only autoregressive generative language models available in different parameter sizes and trained on the Pile dataset \citep{pile}. We selected the following variants for our experiments, in terms of number of model parameters: 410M, 1B, 1.4B, 2.8B, 6.9B and 12B. 

\paragraph{Instruction-tuned language models.} As LLMs readily tuned for instruction-based prompting, we selected Dolly 2.0,\footnote{https://huggingface.co/databricks/dolly-v2-7b} with 7B parameters, which is based on the 6.9B Pythia language model and was tuned on 15K instructions.\footnote{https://huggingface.co/datasets/databricks/databricks-dolly-15k} This model allowed for a direct comparison between the fine-tuned LLM and the instruction-based derived model. Additionally, we selected the Alpaca-Lora model with 7B parameters, based on LLaMA \cite{touvron2023llama} and fined-tuned on the Alpaca instruction dataset,\footnote{https://github.com/gururise/AlpacaDataCleaned} representative of a model derived from a different LLM and set of instructions.

\paragraph{Prompting.} We used different prompting strategies, depending on the model variant. For bare LLMs, we first compare the use of a simple hand-crafted prompt, namely:  


\begin{itemize}[noitemsep]
    \item[] INSTRUCTION: \textit{Split the sentence below into separate sentences.} 
    \item[] INPUT: $\alpha$.
    \item[] RESPONSE: $\beta$
\end{itemize}

\noindent , where $\alpha$ and $\beta$ are provided at training time and the model is tasked to complete $\beta$  at inference time. For this prompting approach, we fine-tuned the Pythia models with LoRA \citep{hu2022lora}, using default parameters (see Appendix~\ref{sec:appendix-params}) . 

Additionally, we experimented with soft-prompting methods, which have been shown to improve LLM performance on several tasks, addressing the limitations of manual prompt design. We selected two such approaches, namely prompt-tuning \cite{lester-etal-2021-power} and P-tuning \cite{liu2023gpt}. In both cases, trainable prompt embeddings are added to a frozen base model, with P-tuning also involving a dedicated encoder.

For zero-shot probing, we used the hand-crafted prompt described above, leaving the response empty for the model to complete. For in-context learning, we used a few-shot approach based on $n$ examples prepended to the prompt. Thus, each query is composed of $n$ concatenated prompts that follow the previously described pattern, along with the last prompt where the response is left empty. For our experiments, we selected the first 5 examples from the development set as examples, although other variants are of course possible.

Preliminary experiments with alternative zero-shot prompt variants led to little variation in results across metrics, with our selected prompt achieving more balanced results across metrics. For few-shot sampling, larger sets of 10 examples provided minor improvements on some metrics, but degradation in terms of compliance. Results with alternative zero-shot and few-shot prompting are provided in Appendix~\ref{sec:appendix-extended-prompting}.


\subsection{Evaluation}

Following \citet{kim-etal-2021-BiSECT}, we used three separate metrics to evaluate our models. SARI \citep{xu-etal-2016-optimizing}, the standard simplification metric as computed by the EASSE toolkit \citep{alva-manchego-etal-2019-easse},\footnote{https://github.com/feralvam/easse} which provides the proportion of n-grams that are added (\textit{add} in what follows), kept (\textit{keep}) or deleted (\textit{del}).  As a second metric, we used the normalised BertScore (hereafter, BESC) \citep{bertscore},\footnote{https://github.com/Tiiiger/bert\_score} which has proved successful to measure meaning preservation. BLEU scores, computed with sacreBLEU \citep{post-2018-call}, are reported as well, as a precision measure of n-gram overlap between hypotheses and references. 

We also included an additional metric for the task to indicate output compliance (CPL), i.e. the percentage of output sequences that include at least two separate sentences, to directly identify unsplit output. This metric was computed with a simple approximation, counting the number of end of sentence punctuation marks in the output sentence period, question and exclamation marks.

We compare our results with those obtained using the BiSECT and ABCD approaches of \citet{kim-etal-2021-BiSECT} and \citet{gao-etal-2021-abcd}, respectively, the state of the art for the selected corpora. For BiSECT, we used their publicly shared output directly, filtering the sentences that were excluded from the \textit{Selected} test set described in Section~\ref{subsec:corpora}, to maintain strict comparability with the other tested methods.\footnote{https://github.com/mounicam/BiSECT/tree/main/outputs/} For ABCD, the output was computed with  the publicly shared models and code,\footnote{https://github.com/serenayj/DeSSE} on the \textit{Selected} test subset, for a fair comparison.


\begin{figure*}[ht]
\centering
\includegraphics[width=0.48\textwidth]{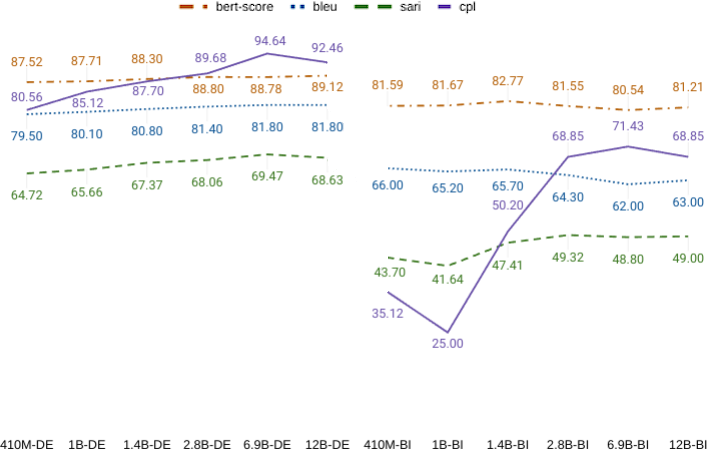}
\hspace*{0.02\textwidth}
\includegraphics[width=0.48\textwidth]{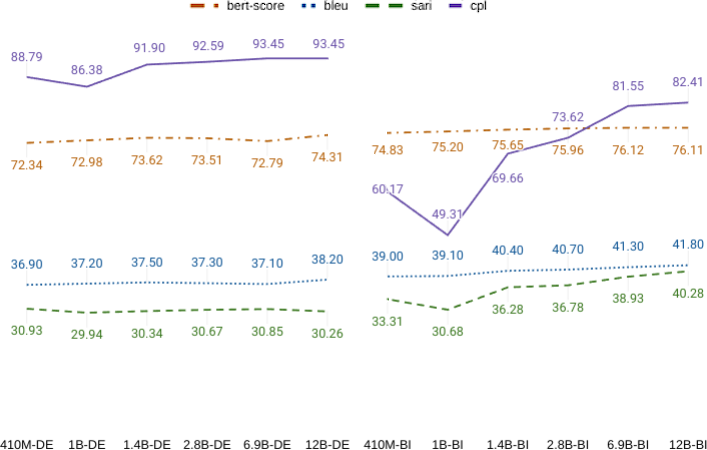}
\caption{Impact of parameter size on the DeSSE (left) and BiSECT (right) test sets, with Pythia model variants fine-tuned over data from the DeSSE (-DE) or BiSECT (-BI) training data.}
\label{fig:llm-sz-com2sim}
\end{figure*}

\begin{figure*}[ht]
\includegraphics[scale=0.35]{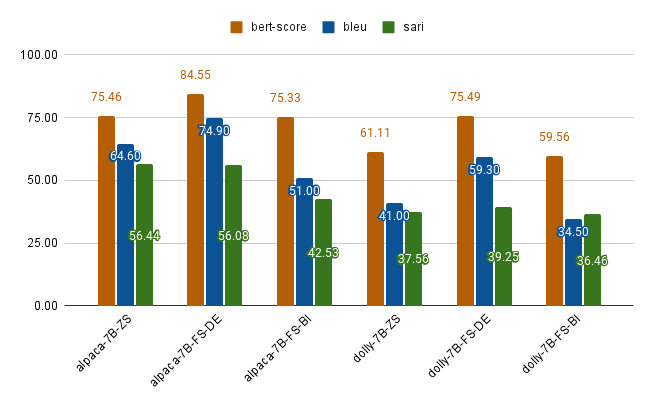}
\includegraphics[scale=0.35]{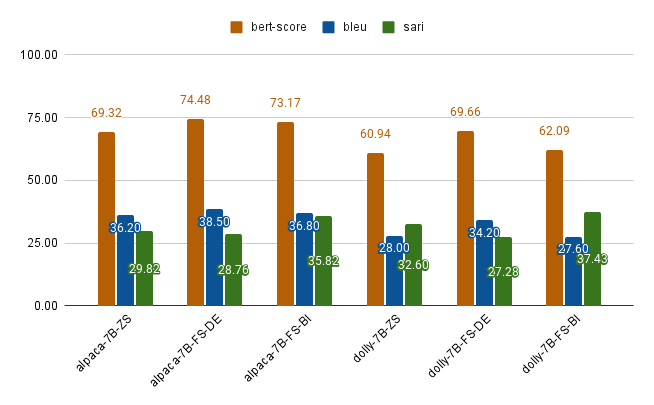}
\caption{Split and Rephrase results on the DeSSE (left) and BiSECT (right) test sets, for instruction-tuned Alpaca and Dolly model variants. Models were applied either in a zero-shot fashion (-ZS), or in a few-shot manner (-FS) with 5 examples sampled from the DeSSE (-DE) or BiSECT (-BI) datasets.}
\label{fig:instruct-llm-com2sim}
\end{figure*}

\section{SPRP with Large Language Models}

\subsection{Prompting Variants}

Table~\ref{table:soft-prompts} shows the results of soft-prompting and LoRA fine-tuning. Excepting two BertScore results, where P-Tuning was marginally better, LoRA adaptation with our hand-crafted simple prompt achieved the best results overall. The differences were relatively small between the three approaches in terms of SARI and BESC, except for PRT-BI on BiSECT with significant losses on all metrics. 



\begin{table}[]
\centering
\begin{tabular}{ccccc}
\toprule
Model & Test &  BESC &  SARI  &  CPL \\ 
\midrule
PRT-DE &  \multirow{3}{*}{\rotatebox{90}{DeSSE}} & 86.81 &  61.02 &   71.03 \\
PTG-DE & & 87.67 &  63.29 &   72.82 \\
LoRA-DE & & \textbf{88.78} &  \textbf{69.47}  &  \textbf{94.64} 
\\ \cmidrule{2-5} 
PRT-DE & \multirow{3}{*}{\rotatebox{90}{BiSECT}} & 73.42 & 27.03 &  63.10 \\
PTG-DE & & \textbf{73.77} &  28.62 &   76.55 \\
LoRA-DE & & 72.79 &  \textbf{30.85} &  \textbf{93.45} \\ 
\midrule
PRT-BI & \multirow{3}{*}{\rotatebox{90}{DeSSE}} & 81.95 & 37.42  & 6.55 \\
PTG-BI & & \textbf{82.00} &  40.20 &  15.67 \\
LoRA-BI & &  80.54 &  \textbf{48.80} &  \textbf{71.43} 
\\ \cmidrule{2-5} 
PRT-BI & \multirow{3}{*}{\rotatebox{90}{BiSECT}} & 37.00 &  1.55 &  8.79 \\
PTG-BI & &  74.71 &  25.32 &  22.07 \\
LoRA-BI & & \textbf{76.12} &  \textbf{38.93} &    \textbf{81.55} \\ 
\bottomrule
\end{tabular}
\caption{Results for Prompt-tuning (PRT), P-Tuning (PTG) and LoRA adaptation with the Pythia-6.9B model trained on DeSSE (-DE) or BiSECT (-BI) data.}
\label{table:soft-prompts}
\end{table}

The most notable result are the significantly larger differences in terms of compliance, where LoRA was markedly better in all cases and both soft-prompting methods failed significantly overall, drastically so when trained on BiSECT data. The poor performance of both methods on the more complex SPRP data is noteworthy, considering the gains achieved with soft-prompting in other downstream tasks. Considering these results, we selected the more robust LoRA approach as our main adaptation method in the remaining experiments. 

\subsection{Fine-tuned LLM Variants}

Figure~\ref{fig:llm-sz-com2sim} shows the results for the Pythia variants fine-tuned on DeSSE (-DE) or BiSECT (-BI).



In terms of parameter size, for in-domain scenarios, on all metrics there is a tendency towards increased accuracy as the number of parameters increases, although only marginally so and with both SARI and CPL actually incurring minor losses with the largest DeSSE-tuned 12B models. Absolute scores were lower with the BiSECT-tuned model on its own test set, which is not unexpected considering the complexity of the test data in this case. CPL and SARI for the BiSECT-tuned model were also more erratic, with a marked drop for the 1B model in particular, which could be due to the challenges posed by training on the complex BiSECT datasets for smaller models. 





Domain shift results provided interesting insights regarding robustness of the variants. For the BiSECT-tuned model, switching from its in-domain test set to the DeSSE test led to higher absolute scores for all metrics but CPL. This is likely due to the lesser complexity of the latter test, although the marked CPL drop could also indicate that training on the more complex BiSECT data resulted in limited portability for the critical splitting part of the task. 

For DeSSE-tuned models, switching to the more complex BiSECT test did result in absolute drops on most metrics, as expected with a significant domain change. However, they achieved comparable results on  CPL despite domains switch, and also achieved comparable results to the BiSECT-tuned model in terms of BESC, BLEU and SARI, while also largely outperformed it on CPL. Modelling SPRP on shorter sentences and limited domain variation, as featured in DeSSE, thus seem to be a better option for higher portability across domains. 



\subsection{Instruction-tuned Models}

The zero-shot and in-context few-shot results with the Alpaca and Dolly models on the SPRP task are shown in Figure~\ref{fig:instruct-llm-com2sim}. The first notable result is the overall higher scores obtained with Alpaca models, on all metrics, when compared to equivalent Dolly variants. The differences were not marginal as they amounted to average differences, between comparable models, of 13.06, 18.57 and 13.93 points in terms of BertScore, BLEU and SARI, respectively, on the DeSSE test set. On the BiSECT test, the average differences between comparable models amounted to 8.09, 7.23	and -0.97, with only the SARI metric leading to comparable results on this test set. The choice of a specific pretrained instruction-tuned LLM for the SPRP task thus seems of particular importance, despite their relative similarity in terms of model architecture, training objectives and dataset variety.

Few-shot variants outperformed their zero-shot alternatives across the board, although it is worth noting that the Alpaca zero-shot model outperformed the Dolly few-shot model overall. As was the case for the fine-tuned Pythia variants, the results on the BiSECT test set were also significantly lower overall than on the DeSSE test, due in part to the higher complexity of this specific test set, which features significantly longer sentences on both the complex and split sides.

\begin{table}[]
\centering
\begin{tabular}{lllrrrrrrr}
\toprule
Train & Size & BESC & BLEU & SARI & CPL \\
\midrule
BI & 1k	& 70.69 & 35.90	&   21.78 & 10.34 \\
BI & 2k	& 74.31 & 36.80 & 21.23 & 5.17 \\ 
BI & 3k	& 75.61	& 40.00 & 32.43 & 52.93 \\
BI & 4k	& 75.82	& 40.70 & 35.69	& 66.03 \\
BI & 6k	&	76.12 & 41.30	& 38.93	& 81.55 \\
BI & 14k	 & 76.04	 & 41.60	 & 41.93	  & 90.69 \\
BI & 30k     & 76.32 &	42.40 &	\textbf{45.06} &	\textbf{95.17} \\
BI & 62k	 & \textbf{76.74} &	\textbf{42.60} &	43.84 &	93.79 \\
\midrule
DE & 1k	 & 69.21	 & 48.60	 & 32.73	 & 13.89 \\
DE & 2k	 & 	 82.26 & 73.90 & 59.99 & 73.21\\
DE & 3k	 & 	 88.25	& 81.10	& 67.33 & 85.52\\
DE & 4k	 & \textbf{88.80}	 & \textbf{81.60} & 68.58	 & 89.09 \\
DE & 6k   & 88.78 & 81.80	&	\textbf{70.39} & \textbf{94.64} \\
\bottomrule
\end{tabular}
\caption{Impact of training data size with the Pythia-6.9B model on BiSECT (BI) and DeSSE (DE)}
\label{table:datasize-results}
\end{table}

\subsection{Impact of Training Data Size}
\label{sec:size}

The results described in the previous sections for the best performing models were obtained with approximately 6,000 training data points. In this section, we aim to evaluate LLM accuracy on the SPRP task with either larger volumes of training data, testing the ceiling of the approach, or smaller amounts to estimate minimal volumes of training data for this method. We used the fine-tuned \textit{Pythia-6.9B} models for these experiments, per the previously discussed rationale. 



The results are shown in Table~\ref{table:datasize-results}. In both cases, scores increased across metrics as training size increased, although by smaller increments from the 3K mark upwards. Best results are split between the two largest training datasets on both BiSECT and DeSSE, indicative of a potential ceiling in terms of improvements from training data augmentation. It might also be the case that the observed plateauing resulted from a lack of variety in the additional training data, but verifying this hypothesis was beyond the scope of this work.

The two domains displayed noteworthy differences. Although using all the available training data resulted in better SARI and CPL scores on DeSSE, those obtained with 3K were closer to the best results than to those obtained with lower amounts. For BiSECT, the largest increase in terms of compliance was achieved at the 6K mark, with a stabilisation above 90\% with training datasets of 14K or more samples. These results are in line with the previously discussed complexity of the BiSECT datasets, compared to DeSSE. It is worth noting that no model variant achieved compliance above 95.2, irrespective of training data size.



\section{Comparative Evaluations}
\label{section:com-eval}

\begin{table*}[]
\centering
\begin{tabular}{llrrrrrrrr}
\toprule
Method & Test & BESC & BLEU & SARI & add & keep & del & CPL \\
\midrule
pythia-6.9B-DE (6K) & DeSSE &  \textbf{88.78} & \textbf{81.80}	&	\textbf{69.47}	& \textbf{ 45.75}	& \textbf{92.26}	& \textbf{70.39} & 94.64\\

alpaca-7B-ZS & DeSSE &  75.46 & 64.60 &	56.44 &	23.14 &	89.35 &	56.82 & 83.93\\
alpaca-7B-FS-DE & DeSSE &  84.55	& 74.90 &	56.08	& 27.76	& 88.92	& 51.56 & 68.65\\

ABCD-bilinear & DeSSE & 66.27 & 57.20 & 52.54 &	17.63	& 81.04	& 58.96 & \textbf{98.41}\\
ABCD-mlp & DeSSE & 78.68 & 67.40	&	57.67	& 22.04	& 87.90	& 63.06 & 89.09\\
\textit{GPT-4 Turbo}  & DeSSE & \textit{84.59} & \textit{70.90} & \textit{63.69} & \textit{35.24} & \textit{89.76} & \textit{66.09} & \textit{99.20} \\
\midrule

pythia-6.9B-BI (6K) & BiSECT  & 76.12	& 41.30	& 38.93 &	12.92 &	63.59 &	40.27 & 81.55\\

pythia-6.9B-BI (30k) &  BiSECT &   \textbf{76.32} &	\textbf{42.40} &	\textbf{45.06} & \textbf{16.81} & \textbf{65.93} & \textbf{52.45} & 95.17 \\

alpaca-7B-ZS &  BiSECT & 69.32 & 36.20	& 29.82	& 5.08 &	60.39 &	23.99 & 88.62\\
alpaca-7B-FS-BI &  BiSECT & 73.17 &	  36.80 &	35.82	& 8.57 &	60.46 &	38.43 & 65.86\\

BiSECT seq2seq &  BiSECT & 72.51	& 36.80 & 33.21 &	7.34	& 59.91	& 32.40 & \textbf{99.48}\\
\textit{GPT-4 Turbo}  & BiSECT & \textit{72.04} & \textit{34.70} & \textit{37.18} & \textit{9.65} & \textit{60.33} & \textit{41.55} & \textit{100.00} \\

\bottomrule
\end{tabular}
\caption{Comparative metrics results for the Split and Rephrase task on the DeSSE and BiSECT test sets.}
\label{table:comp-results}
\end{table*}

\subsection{Metrics Results}



From the results in the previous sections, we selected a subset of models for comparison with the ABCD and BiSECT approaches, namely: the best 6.9B Pythia models in terms of CPL and SARI (Section~\ref{sec:size}), which obtained only slightly lower results on all metrics than the 12B variant and features a comparable amount of parameters to the instruction-tuned LLMs; the Alpaca zero-shot model, representative of scenarios where no data are available for fine-tuning or in-learning; and the Alpaca few-shot model using either BiSECT and DeSSE examples. For the ABCD approach, we used both the bilinear and MLP variants. 

We also included results from GPT-4 Turbo\footnote{Available at https://help.openai.com/en/articles/8555510-gpt-4-turbo. Version \textit{gpt-4-1106-preview} February 2024. We used the same instruction as for our other zero-shot results.}, as one of the largest LLMs to date, although these results should be considered with caution: training data are unknown, with possible test data leakage, and  reproducibility standards are not met, considering the proprietary nature of the model. The comparative results are shown in Table~\ref{table:comp-results}.


On both test sets, the fine-tuned Pythia models with 6.9B parameters outperformed all alternatives. On DeSSE, the best ABCD model (ABCD-mlp) trailed by 10.1, 14.4, 11.8 and 5.55 points in terms of BESC, BLEU, SARI and CPL, respectively; the ABCD-bilinear model obtained the best compliance score but the lowest results on the other metrics. On BiSECT, the BiSECT model was outperformed by 3.81, 5.6 and 11.85 points on  BESC, BLEU and SARI, respectively; it was however the most compliant model. The LLM variants can thus provide more accurate results across the board, but can still lag behind alternatives in terms of enforcing split SPRP hypotheses.


It is worth noting that instruction-tuned models prompted with only 5 in-context examples obtained competitive results across the board, although with significantly lower compliance scores overall. The zero-shot results were also competitive against the ABCD and BiSECT approaches, although they obtained the lowest scores overall. The fine-tuned Pythia models were thus shown to be a better alternative when training datasets of sufficient size are available, but the instruction-based models could provide a reasonable alternative.



Finally, although GPT-4 Turbo achieved the highest compliance, it was outperformed by our best LoRA variants on all other metrics. After manual examination of sampled output, the main reason seems to be a stronger tendency to rephrase and extend the original content. Different prompts might mitigate this issue, an option we left aside to maintain a fair comparison between models.

\subsection{Manual Ranking Results}

Metrics such as BLEU have important limitations for the SPRP task, displaying low correlation with human judgements in some experiments \cite{zhang-etal-2020-small}. Although the results of the previous section show large improvements across multiple metrics, indicative of marked improvements achieved by LLM-based SPRP, we conducted an additional manual comparative evaluation. For this specific evaluation, human evaluators were presented a source sentence and two SPRP hypotheses, which they had to rank in a 3-way evaluation, with either one of the hypotheses considered better, or both considered equally correct or incorrect. 

For this task, we adapted the Appraise environment \cite{federmann2018appraise}, typically used for machine translation evaluation campaigns, and asked 11 evaluators to perform two sub-tasks, each involving 100 samples randomly selected: compare hypotheses from fine-tuned Pythia-6.9B models with those of ABCD-mlp and BiSECT models, on the DeSSE and BiSECT test sets, respectively. Evaluators were asked to compare the hypotheses according to SPRP criteria of compliance, sensicality, grammatically, missed facts and omitted facts.\footnote{See Appendix~\ref{sec:appendix-3wayeval} for details on the experimental protocol.} The results are shown in Table~\ref{table:3way-results}.

\begin{table}
\centering
 \begin{tabular}{ l c c c  c }
 \toprule
S & P > S & P = S & P < S & $\alpha$\\
 \midrule
 ABCD & 70.27\%  & 28.55\% & 1.18\% & 0.52\\
 BiSECT & 43.09\% & 38.45\% & 18.45\% & 0.49 \\
 \bottomrule 
 \end{tabular}
 \caption{3-way ranking results comparing hypotheses from Pythia-6.9B-DE (P) and alternative SPRP methods (S) ABCD or BiSECT, on the DeSSE and BiSECT test sets, respectively. $\alpha$ indicates Krippendorf's inter-annotator agreement.}
 \label{table:3way-results} 
\end{table}

Compared to ABCD, LLM-based hypotheses were largely preferred by human evaluators, at 70.27\%, with 28.55\% considered equal. On the more complex BiSECT test data, which mainly featured long sentences in the pharmaceutical and medical domain, the results were more balanced, with 38.45\% of hypotheses considered equally good or bad. However, LLM hypotheses were still markedly preferred for the remaining hypotheses, at 43.09\% against 18.45\% of hypotheses selected from the BiSECT sequence-to-sequence model. In both evaluations, inter-annotator agreement was moderate overall, at 0.52 and 0.49, slightly lower on the more complex BiSECT data.

The results obtained in this comparative human evaluation are thus in line with those obtained on automated metrics overall, with marked improvements achieved via LLM-based SPRP.

\section{Qualitative Evaluation}
\label{section:qual-eval}

To gain further insights on the SPRP capabilities of LLMs, we performed an additional human evaluation, this time centred on assessing the validity of individual SPRP hypotheses on a set of predefined criteria. For this qualitative evaluation, we sampled 100 source sentences from Wiki-BM \cite{zhang-etal-2020-small}, a manually crafted benchmark created from Wikipedia data, and generated SPRP hypotheses with the Pythia-6.9B model fine-tuned on DeSSE data. The evaluation involved 8 human evaluators, who had to answer 6 questions related to the quality of an SPRP hypothesis considering a source sentence, addressing sencicality, grammaticality, deletion or introduction of new facts, correct splits and insufficient splits.\footnote{The questions were adapted from \citet{zhang-etal-2020-small}, with minor modifications. See Appendix~\ref{sec:appendix-qualeval} for details on the experimental protocol.} 




Overall, the results, shown in Table~\ref{table:qual-results}, were largely positive. Most output was judged sensical, grammatical, with almost no introduction of new facts, and correct splits in over 91\% of the cases. All were judged with almost perfect or strong inter-annotator agreement, with sensicality and correct splitting slightly below at around 80\% agreement. 

Although still featuring scores above 80\%, the remaining aspects led to comparatively lower marks, with missed facts and not enough splits in  15.87\% and 16.75\% of the cases, respectively, and inter-rater agreement between 0.55 and 0.60. Missed facts in particular were somewhat more difficult to judge, according to post-mortem feedback, as some rephrasing cases involved implicit rather than explicit logical connection between sentences, which could be rated more or less strictly. Similarly, the need for additional splits can be viewed as more subjective than other categories, as evaluators could view some grammatically correct additional splits as cases of over-splitting.\footnote{Examples of LLM output from this evaluation are shown in Appendix~\ref{sec:appendix-examples}. Metrics results on the complete Wiki-BM benchmark are indicated in Appendix~\ref{sec:appendix-wikibm-metrics}.} 

It is worth noting that these results were obtained with a model fine-tuned on data from a different domain than Wiki-BM, namely DeSSE, showing that knowledge extracted from one dataset could be effectively transferred to other domains.

\begin{table}
\centering
 \begin{tabular}{ l c c c  c }
 \toprule
 Question &  Answer & $\alpha$ \\
 \midrule
Sensical? [0-2] & 86.12\% / 1.84 & 0.803\\
Grammatical? [0-2] & 92.62\% / 1.91 & 0.907 \\
No missed facts? [Y/N]  & 84.13\% & 0.597 \\
No new facts? [Y/N] & 99.50\% & 0.996\\
Correct split? [Y/N] & 91.13\% & 0.797 \\
Enough split? [Y/N] & 83.25\% & 0.548 \\
 \bottomrule 
 \end{tabular}
 \caption{Qualitative evaluation results with Pythia-6.9B on Wiki-BM ($\alpha$ = Krippendorf's alpha inter-rater agreement). Answers for the first two questions are the percentage of 2 and the average score on a [0-2] scale. For the other questions, we report the percentage of Yes.}
 \label{table:qual-results} 
\end{table}





\section{Conclusions}

In this work, we evaluated the performance of large language models on the Split and Rephrase task, improving over the state of the art by significant margins on all metrics over SPRP datasets of varying complexity. The results of a manual 3-way ranking evaluation further confirmed human preference for LLM-based hypotheses, and a qualitative evaluation on Wiki-BM data further demonstrated the quality of the generated SPRP output.

Among LLM variants, models fine-tuned with LoRA achieved the best results overall. Zero-short and few-shot instruction-tuned variants achieved reasonable performance, modulo a significant drop in splitting compliance for the latter, and might thus constitute a reasonable alternative in languages where training corpora are lacking for the task. 

Increasing fine-tuning data size resulted in only gradual improvements, plateauing on most metrics with subsets of the available data. Increased parameters led to only minor improvements, with models around 7B achieving the more balanced results. Despite significant improvements, none of the examined variants achieved full task compliance, and core aspects of SPRP remain a challenge.

\section{Limitations}

The first limitation of an LLM-based approach is the need for pretrained LLMs for the specific languages in which the SPRP task needs to be performed. Note however that even English-centric LLMs can exhibit multilingual capability \cite{armengol-estape-etal-2022-multilingual}, thus mitigating the issue. Additionally, multilingual models such as BLOOM \cite{scao2022bloom} can provide support for the task; we provide preliminary results in this sense in Appendix~\ref{appendix:multilingual}. A more in-depth evaluation of LLM performance on the SPRP task, for languages with limited coverage, would be required to further assess the limitations of the approach.

A second limitation concerns the dependency on training data to fine-tune pretrained LLMs. We established that around 3K pairs of complex and split sentences might be necessary for the task, with high quality results obtained using between 4K and 6K training pairs, depending on the domain. Although this type of training data can be collected from either parallel multilingual datasets, as in BiSECT, or specific types of essays, as in DeSSE, this still requires dedicated work for new languages. Machine translating existing SPRP corpora with high quality models could provide a basis for fine-tuning in other languages, although a precise evaluation of this approach would be needed (see Appendix~\ref{appendix:multilingual} for preliminary results along these lines). We also showed that instruction-tuned pretrained models could provide quality results using only 5 in-context learning pairs, which could be rapidly crafted manually.

\section{Ethical Considerations}
Our approach involves the use of large language models, whose computational performance is typically higher when deployed in more powerful environments with GPUs, for either training or inference. Under such usage, electric consumption and associated carbon footprint are likely to increase and users of our method under these conditions should be aware of this type of impact. Our approach mitigates these costs via the use of pretrained language models of relatively limited size, and an efficient tuning method such as LoRA \cite{hu2022lora} which minimises training time and computational resources.

\section*{Acknowledgements}

We wish to thank the anonymous ARR reviewers for their helpful comments, and the participants in our manual evaluations for their time and contribution. This work was  partially supported by the Department of Economic Development and Competitiveness of the Basque Government (Spri Group) through funding for the IRAZ project (ZL-2024/00570).

\bibliography{anthology,custom}

\onecolumn
\newpage
\appendix

\section{LLM Training and Inference}
\label{sec:appendix-params}


Fine-tuning was performed with LoRA \cite{hu2022lora} using the following default parameters: r=8, alpha=16, dropout=0.05; targeted modules were: Query, Key and Value. We also used default generation parameters, namely: temperature=0.1; top\_p=0.75; top\_k=40; num\_beams=4; max\_new\_tokens=128.

Each training sample was provided in the following format, where $\alpha$ is a complex sentence and $\beta$ its corresponding split sentences, sampled from the DeSSE and BiSECT training sets:

\begin{itemize}

    \itemsep0em 

    \item[] INSTRUCTION: Split the sentence below into separate sentences.

    \item[] INPUT: $\alpha$ 

    \item[] RESPONSE: $\beta$    

\end{itemize}

The following text was prepended to each sample prompt: "\textit{Below is an instruction that describes a task, paired with an input that provides further context. Write a response that appropriately completes the request.}" 

Each training process was performed on 1 A40 GPU with 48GB of RAM. As reference times, fine-tuning took 06:07:59 hours for the Pythia-6.9B model on the BiSECT 6K training dataset and 48 hours over the larger 64K dataset. Fine-tuning the Pythia 12B model took 06:07:59 hours and the smaller 410M model took 00:37:51 hours, on the 6K BiSECT dataset.

For soft-prompting methods, we used the following default settings: num\_virtual\_tokens=20, prompt\_tuning\_init="\textit{Split the sentence below into separate sentences.}" for Prompt-tuning and num\_virtual\_tokens=20, encoder\_hidden\_size=128 for P-tuning.
    
\section{Split and Rephrase Examples}
\label{sec:appendix-examples}

In Tables~\ref{table:desse_examples}, \ref{table:bisect_examples} and \ref{table:wikibm_examples}, we provide illustrative examples of split and rephrased output obtained with the fine-funed Pythia-6.9B model on Desse, BiSECT and Wiki-BM source sentences, respectively. We selected examples that are representative of the type of alterations to the source performed by the selected model, which was used for the manual evaluations, including errors in splitting and/or rephrasing.

Sequences from the source text that have been correctly deleted or modified are shown in purple. Sequences that have been correctly added to the output are shown in green. Sequences that have been incorrectly altered in the original sentence are shown in red.

\begin{table*}
\begin{tabular}{p{7.5cm}|p{7.5cm}}
\hline
\multicolumn{1}{c}{\textbf{Original Text}} &
  \multicolumn{1}{c}{\textbf{Split \& Rephrased Text}} \\ \hline

Most people travel to Europe or the Islands \colorbox{blue!30}{and} get a taste of the culture in those places. &
  Most people travel to Europe or the Islands\colorbox{green!30}{. They} get a taste of the culture in those places. \\ \hline
People would've been killed\colorbox{blue!30}{,} places would have been bombed \colorbox{blue!30}{and} there would have been so much going on all at once. &
  People would've been killed\colorbox{green!30}{.} Places would have been bombed\colorbox{green!30}{.} There would have been so much going on all at once. \\ \hline
War is injuring innocent people \colorbox{blue!30}{which} is not necessary. &
  War is injuring innocent people\colorbox{green!30}{. It} is not necessary. \\ \hline
As education continues to reach the poorer areas of our world i truly believe that one day we will all live in TRUE harmony. &
  Education continues to reach the poorer areas of our world\colorbox{green!30}{.} I truly believe that one day we will all live in TRUE harmony. \\ \hline
This land is still the home to many civilians \colorbox{blue!30}{who} rely on the resources in their area to live. &
  This land is still the home to many civilians\colorbox{green!30}{. They} rely on the resources in their area to live. \\ \hline
\end{tabular}
\caption{SPRP examples on the DeSSE dataset}
\label{table:desse_examples}  
\end{table*}

\begin{table*}
\begin{tabular}{p{7.5cm}|p{7.5cm}}
\hline
\multicolumn{1}{c}{\textbf{Original Text}} &
  \multicolumn{1}{c}{\textbf{Split \& Rephrased Text}} \\ \hline
  
Dialysis does not \colorbox{blue!30}{alter serum doxycycline} \colorbox{blue!30}{half-life and} thus would not be of benefit in \colorbox{blue!30}{treating} cases of overdose. &
  Dialysis does \colorbox{green!30}{not change the serum half-life of} \colorbox{green!30}{doxycycline.} Thus\colorbox{green!30}{,} it would not be of benefit in the treatment of cases of overdose. \\ \hline
In the first study, type \colorbox{red!30}{A resistant patients} \colorbox{blue!30}{(}A-non responders\colorbox{blue!30}{)} were randomised to receive placebo or 10, 000 U of NeuroBloc \colorbox{blue!30}{and} in the second, type A toxin responsive patients \colorbox{red!30}{(A-responders)} were randomised to receive placebo, 5000 U or 10, 000 U of toxin. &
  In the first study, type \colorbox{green!30}{A non-responders} were randomized to receive placebo or 10, 000 U of NeuroBloc\colorbox{green!30}{.} In the second \colorbox{green!30}{study}, type A responders were randomized to receive placebo, 5000 U or 10, 000 U of toxin. \\ \hline
Most patients treated with Zevalin experience a \colorbox{blue!30}{severely reduced} number of blood cells\colorbox{blue!30}{, which} needs some time to \colorbox{blue!30}{normalise again} after treatment. &
  Most patients treated with Zevalin experience a \colorbox{green!30}{severe reduction in the} number of blood cells\colorbox{green!30}{. It} takes some time \colorbox{green!30}{to return to normal} after treatment. \\ \hline
Lisonorm tablet contains 5mg amlodipine\colorbox{blue!30}{/} 10mg lisinopril \colorbox{blue!30}{and is an anti-hypertensive indicated} in patients \colorbox{blue!30}{with} blood pressure adequately controlled with lisinopril and amlodipine \colorbox{red!30}{given concurrently at the same dose level}. &
  Lisonorm tablet contains 5 mg of amlodipine \colorbox{green!30}{and 10 mg of} lisinopril\colorbox{green!30}{. It is} indicated \colorbox{green!30}{for the treatment of hypertension} in patients \colorbox{green!30}{whose blood pressure is} adequately controlled with lisinopril and amlodipine. \\ \hline

\end{tabular}
\caption{SPRP examples on the BiSECT dataset}
\label{table:bisect_examples}  
\end{table*}

\begin{table*}
\begin{tabular}{p{7.5cm}|p{7.5cm}}
\hline
\multicolumn{1}{c}{\textbf{Original Text}} &
  \multicolumn{1}{c}{\textbf{Split \& Rephrased Text}} \\ \hline
  

She walked out on their wedding \colorbox{blue!30}{, and} she currently is single \colorbox{blue!30}{and} holds intense anger towards Richard for ruining their wedding, \colorbox{blue!30}{and} \colorbox{red!30}{also} became furious to learn she is pregnant, \colorbox{red!30}{even though} Richard told her he got a vasectomy.	& She walked out on their wedding\colorbox{green!30}{. }She currently is single\colorbox{green!30}{. She} holds intense anger towards Richard for ruining their wedding\colorbox{green!30}{. She} became furious to learn she is pregnant\colorbox{green!30}{.} Richard told her he got a vasectomy. \\ \hline

Southfort is an unincorporated community in DeSoto County, Florida, United States\colorbox{blue!30}{,} located approximately south of Fort Ogden on U.S. Route 17.	& Southfort is an unincorporated community in DeSoto County, Florida\colorbox{green!30}{. It is} located approximately south of Fort Ogden on U.S. Route 17. \\ \hline

Parke played for powerhouse youth club FC Delco\colorbox{blue!30}{, and} played four years of college soccer at Drexel University\colorbox{blue!30}{,} \colorbox{red!30}{before} being drafted by the MetroStars with 60th and last overall pick of the 2004 MLS SuperDraft.	& Parke played for powerhouse youth club FC Delco\colorbox{green!30}{. He} played four years of college soccer at Drexel University\colorbox{green!30}{. He} was drafted by the MetroStars with 60th and last overall pick of the 2004 MLS SuperDraft. \\ \hline

Auguste Le Duc started on a bridge between the two islands\colorbox{blue!30}{, which} was quickly swept away by the forces of nature.	& Auguste Le Duc started on a bridge between the two islands\colorbox{green!30}{. The bridge} was quickly swept away by the forces of nature. \\ \hline

\end{tabular}
\caption{SPRP examples on the Wiki-BM dataset}
\label{table:wikibm_examples}  
\end{table*}
  



\section{Extended Prompting Variants}
\label{sec:appendix-extended-prompting}

\paragraph{Few-shot prompts.} As noted in Section \ref{sec:models}, we explored the use of larger few-shot sampling on the instruction-tuned models, using 10 in-context learning examples instead of the 5 selected for our core experiments. Results of these additional experiments are presented in Table~\ref{table:few-shot-alpaca} for the best-performing instruction-tuned Alpaca 7B model.

Prompting with 10 examples led to relatively minor gains in SARI in most cases, with a marked tendency to delete more content in the rephrased version. BertScore and BLEU were virtually similar. Extending the in-context samples also resulted in CPL losses with domain shift, up to 10 points with 10-shot--BI on the DeSSE test set, notably, although the reverse was observed with 10-shot-DE on the BiSECT test, with a 2.59 points gain. A minor CPL loss was also observed with 10-shot-BI over BiSECT. 

Domain shift losses were not totally surprising considering the in-context conditioning on samples from the very different DeSSE and BiSECT domains. Considering these results, in particular CPL losses, we selected 5-shot as our primary method for few-shot in-context learning.

\paragraph{Zero-shot prompts.} The zero-shot prompt selected for our experiments was chosen after preliminary testing of different manually engineered variants. No clear better prompting candidate emerged from this initial phase, although it is always possible that an alternative prompt could lead to better result with a given model. To provide further information on zero-shot prompting variants, in Table~\ref{table:prompt-variants} we report additional results with the following prompts, against GPT-4 Turbo (\textit{gpt-4-1106-preview}, accessed February 2024):

\begin{itemize}
   \item Prompt O:  "Split the sentence below into separate sentences."
   \item Prompt A:"Split the following sentence into multiple shorter sentences, while preserving the original meaning as closely as possible."
   \item Prompt B:"Rephrase the given sentence to make it more concise and direct, without changing the core meaning."
   \item Prompt C:"Split the following sentence into multiple concise sentences, separating each distinct idea."
\end{itemize}

Prompt O is our original prompt, used throughout our experiments, and prompts A-C are alternative manually engineered prompts for the task.\footnote{We thank the anonymous ARR reviewer who suggested these specific prompt variants as additional test cases.}
Leaving aside prompt B, which resulted in almost no compliance and severe degradation overall, the other two prompt variants achieved relatively similar results compared to Prompt O. Overall, the latter still achieved better results on all metrics on DeSSE, and all but SARI on BiSECT. All prompt variants were outperformed by the best fine-tuned 6.9B variants, except on compliance.

\begin{table}[]
\centering
\begin{tabular}{ccccccccc}
\toprule
Test          & Method & BESC  & BLEU  & SARI  & add   & keep  & del   & CPL   \\ \hline
\multirow{4}{*}{DeSSE} & 5-shot-DE   & 92.51          & 74.90          & 56.08          & 27.76          & \textbf{88.92}          & 51.56          & 68.65          \\
                       & 10-shot-DE   & \textbf{92.75} & \textbf{75.10} & \textbf{58.07} & \textbf{31.49} & 88.46          & \textbf{54.26}          & \textbf{71.83}          \\ 
                        & 5-shot-BI  & 88.04          & 51.00          & 42.53          & 11.51          & 73.53          & 42.55          & 31.94          \\
                        & 10-shot-BI & 87.40          & 47.90          & 40.09         & 8.55           & 71.44          & 40.28          & 21.63          \\ \midrule
\multirow{4}{*}{BiSECT} & 5-shot-BI  & 86.99          & 36.80          & 35.82          & 8.57           & \textbf{60.46} & 38.43          & 65.86          \\
                        & 10-shot-BI & 86.60          & 35.70          & \textbf{38.06} & \textbf{10.08} & 60.03          & \textbf{44.09} & 64.31          \\
                        & 5-shot-DE   & 87.63          & \textbf{38.50} & 28.76          & 5.46           & 59.92          & 20.89          & 66.38          \\
                        & 10-shot-DE  & \textbf{87.73} & 38.00          & 29.32          & 5.74           & 59.62          & 22.60          & \textbf{68.97}          \\                         
\bottomrule
\end{tabular}
\caption{Extended few-shot results with the Alpaca-7B model using examples from DeSSE (DE) and BiSECT (BI).}
\label{table:few-shot-alpaca}
\end{table}

\begin{table}[]
\centering
\begin{tabular}{lllllllll}
\toprule
Method      & Test & BESC  & BLEU & SARI  & add   & keep  & del   & CPL \\ \hline
GPT-4 Prompt O & DeSSE         & \textbf{84.59} & \textbf{70.90} & \textbf{63.69} & \textbf{35.24} & \textbf{89.76} & \textbf{66.09} & 99.20         \\
GPT-4 Prompt A & DeSSE  & 79.92          & 59.40          & 56.52          & 25.74          & 82.94          & 60.86          & 99.80         \\
GPT-4 Prompt B & DeSSE  & 61.87          & 15.40          & 26.27          & 2.58           & 37.18          & 39.05          & 1.38         \\
GPT-4 Prompt C & DeSSE  & 79.27          & 58.50          & 55.45          & 24.24          & 81.89          & 60.24          & \textbf{100.00} \\
\textit{pythia-6.9B-DE (6K)} & DeSSE         & \textit{88.78 }& \textit{81.80} & \textit{69.47} & \textit{45.75} & \textit{92.26} & \textit{70.39} & \textit{94.64}         \\
\midrule
GPT-4 Prompt O & BiSECT & \textbf{72.04} & \textbf{34.70} & 37.18          & 9.65           & 60.33          & 41.55          & \textbf{100.00} \\
GPT-4 Prompt A & BiSECT & 67.78          & 29.30          & 42.54          & 10.36          & 59.92          & \textbf{57.33} & \textbf{100.00} \\
GPT-4 Prompt B & BiSECT & 65.37          & 18.40          & \textbf{43.38} & 9.04           & 49.90           & 71.20           & 3.96         \\
GPT-4 Prompt C & BiSECT & 68.14          & 30.60          & 42.59          & \textbf{10.46} & \textbf{60.61} & 56.69          & \textbf{100.00} \\
\textit{pythia-6.9B-BI (30K)} & BiSECT         & \textit{76.32}& \textit{42.40} & \textit{45.06} & \textit{16.81} & \textit{65.93} & \textit{52.45} & \textit{95.17}         \\
\bottomrule
\end{tabular}

\caption{Zero-shot prompt variant results on DeSSE and BiSECT with GPT4. Pythia-6.9B results are provided as a reference. Best results amongst GPT-4 variants are shown in bold. }
\label{table:prompt-variants}
\end{table}

\section{Evaluation Metrics on Wiki-BM}
\label{sec:appendix-wikibm-metrics}

In section~\ref{section:qual-eval}, we used portions of the Wiki-BM corpus \citep{zhang-etal-2020-small} for our qualitative evaluation of LLM-based SPRP output. In Table~\ref{table:wikibm-metrics}, we provide the results in terms of metrics for the complete Wiki-BM test set, along with results from \citet{alajlouni2023knowledge}, as representative of an alternative method based on fine-tuning an encoder-decoder pretrained T5 model, and GPT4 Turbo, as one of the largest LLMs to date. The caveats mentioned in Section~\ref{section:com-eval} regarding the latter apply to these results as well.

The results for our LLM-based approach are in line with those obtained on the DeSSE test set across metrics: slightly lower on BESC, BLEU, and SARI, but higher in terms of compliance (CPL). The Pythia-6.9B model fine-tuned with LoRA notably outperformed the reported scores in \citet{alajlouni2023knowledge} by 13.4 and 5.1 points on the BLEU and SARI metrics, respectively. GPT-4 Turbo achieved the best results on all metrics, but by relatively small margins overall. These results confirm the viability of relatively smaller LLMs, tuned with LoRA on small amounts of data, for the SPRP task.


\begin{table*}[]
\centering
\begin{tabular}{lrrrrrrrr}
\toprule
Method & BLEU & SARI & add & keep & del & BESC & CPL \\
\midrule

pythia-6.9B-DE	& 72.90	& 65.10	& 36.14	& 91.15	& 68.10	& 85.50	  & 98.19 \\

T$_{MWS}$ \cite{alajlouni2023knowledge} & 59.50 & 60.00 & - & - & - & - & -  \\ 
\textit{GPT-4 Turbo} & \textit{74.40} & \textit{68.63} & \textit{39.96} & \textit{91.90} & \textit{74.03} & \textit{86.23} & \textit{99.70} \\


\bottomrule
\end{tabular}
\caption{Metrics results on the Wiki-BM benchmark test set. Comparison between Pythia-6.9B fine-tuned on DeSSE data, results reported in \citep{alajlouni2023knowledge}, and GPT4 Turbo (accessed on February 2024).}
\label{table:wikibm-metrics}
\end{table*}

\section{3-Way Ranking Evaluation Protocol}
\label{sec:appendix-3wayeval}

\begin{figure*}
\centering
\includegraphics[scale=0.5]{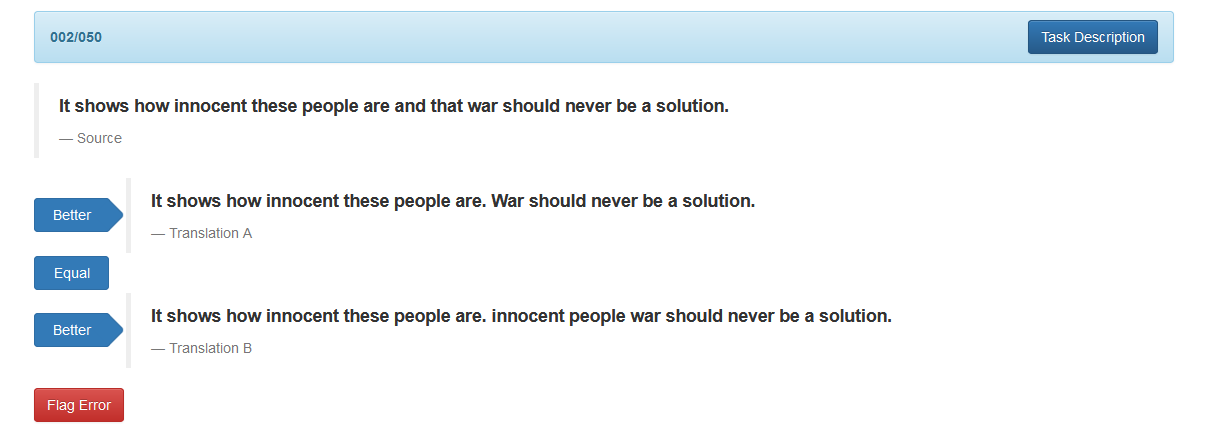}
\caption{Appraise interface adapted for the 3-way ranking SPRP evaluation}
\label{fig:appraise-eval}
\end{figure*}

The 3-way ranking manual evaluation involved 11 human evaluators, all researchers with at least upper-intermediate knowledge of English, who were not paid for the task. The evaluation was split into two tasks: for one task, the evaluators were asked to rank the output of the fine-tuned Pythia-6.9B model and the output of the ABCD model; for the other task, they had to rank the output of the fine-tuned Pythia-6.9B model and the output of the BiSECT model. Each of the two tasks involved 100 samples. Evaluators were also asked to ignore minor capitalisation and tokenisation issues, which were typically present in the output of the ABCD and BiSECT models, as detected in preliminary data inspection.

We adapted the Appraise framework \cite{federmann2018appraise} for the task, mainly by simplifying the presentation of the options, and the following instructions were provided to complete the task:

\begin{itemize}

    \item The task is named Split and Rephrase (SPRP) and consists in splitting a single complex sentence into a sequence of two or more grammatical sentences and rephrasing them as needed to preserve the original meaning without omitting or adding information.

    \item You will be shown 100 examples for each task, each example as shown in Figure~\ref{fig:appraise-eval}, where:

    \begin{itemize}

        \item  The sentence shown at the top (source) is the original single sentence. 

        \item The sentences shown as translation A and translation B are SPRP hypothesis which you need to compare.

        \item After reading the source sentence and each SPRP hypothesis, you should perform one of the following actions:

        \begin{itemize}
            \item Click on the \textit{Better} button next to one of the hypothesis if it is better than the other hypothesis according to the criteria below;
            
            \item Click on the \textit{Equal} button if the two hypothesis are both equally correct or both equally incorrect, according to the criteria indicated below.
        \end{itemize}
        
    \end{itemize}

    \item An SPRP hypothesis can be considered correct if:

    \begin{itemize}
        \item it contains at least two separate sentences (\textit{compliance}),
        \item it contains all the information in the source sentence (\textit{no missing facts}),
        \item it does not contain information that was not present in the source sentence (\textit{no new facts}),
        \item all the sentences in the hypothesis are grammatical (\textit{grammaticality}),
        \item all the sentences in the hypothesis make sense both in isolation and in sequence (\textit{sensicality}).        
    \end{itemize}    

    \item Ranking criteria:

    \begin{itemize}

        \item If both hypotheses meet all criteria, select \textit{Equal}.
        \item If neither hypothesis meets any of the criteria, select \textit{Equal}.
        \item If one hypothesis meets all criteria and the other does not, select the former as \textit{Better}.
        \item If one of the hypotheses is markedly better than the other, in the sense that as it meets almost all criteria whereas the other fails to meet most criteria, and only contains minor errors that do not impede comprehension of the original information, you may select the former as \textit{Better}. In case of doubt, select \textit{Equal}.
        \item When validating an hypothesis, please ignore minor capitalisation errors (e.g., at the beginning of a sentence) and minor tokenisation errors (e.g., punctuation preceded by a space).\footnote{This advice was meant to avoid discarding the ABCD or BiSECT hypotheses on these non-SPRP criteria, as they were shown to contain this type of errors in their respective output, upon preliminary data inspection.}
        
    \end{itemize}
        
\end{itemize}

System hypotheses are presented in random order by default in the Appraise environmnent.

\section{Qualitative Evaluation Protocol}
\label{sec:appendix-qualeval}

\begin{figure*}
\centering
\includegraphics[scale=0.5]{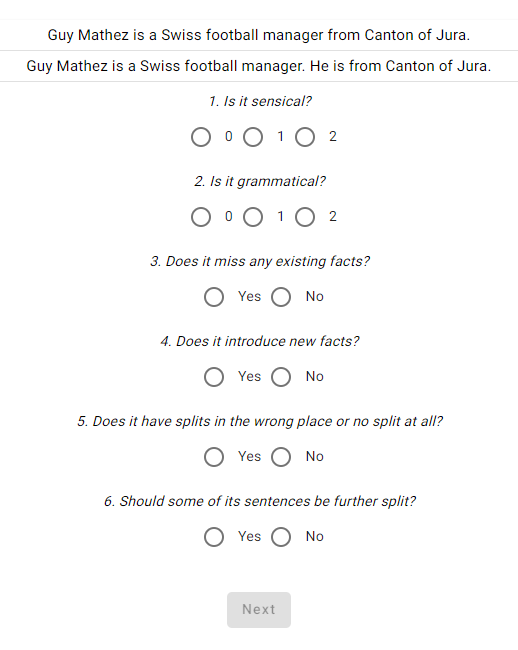}
\caption{Web interface for the qualitative evaluation}
\label{fig:qualweb-eval}
\end{figure*}

The evaluation involved 8 human evaluators, all researchers with at least upper-intermediate knowledge of English, who were not paid for the task. Each evaluator had to answer 6 questions related to the quality of an SPRP hypothesis considering a source sentence. We randomly sampled 100 sentences from the Wiki-BM corpus \cite{zhang-etal-2020-small} and generated SPRP hypotheses with the Pythia-6.9B model fine-tuned on DeSSE data.

The questions were taken from \citet{zhang-etal-2020-small} for the most part, with minor modifications: a simplified 0-2 Likert scale instead of their 0-5 scale for the first two questions, as the latter appeared more difficult to use in practice, and a reformulation of question 5 to explicitly include cases where the output was not split.

The samples and associated questions were provided within an ad-hoc Web-based environment developed for the task. The following instructions were provided to the evaluators:

\begin{itemize}

    \item Upon entering the Web-based evaluation environment, you will be shown a page as in Figure~\ref{fig:qualweb-eval}.

    \item The topmost sentence is the original source sentence. The sentence below is a Split \& Rephrase (SPRP) hypothesis, i.e., a rewrite of the source sentence into a sequence of separate sentences that preserve the meaning of the source sentence. 

    \item For each pair of source sentence and SPRP hypothesis, you will need to answer all 6 questions in the page. The guidelines for each question are indicated below:

    \begin{itemize}
    
        \item \textbf{Is it sensical? (scale of 0-2)} 
            \begin{itemize}    
                \item Indicate:         
                    \begin{itemize}
                        \item 0 if the SPRP hypothesis contains major sensicality errors (the text is barely comprehensible); 
                        \item 1 if it contains minor sensicality errors (the text is comprehensible but has lost part of the original meaning, e.g., lacking logical connection between sentences); 
                        \item 2 if there are no sensicality errors.                        
                     \end{itemize}        
                \item Examples:
                    \begin{itemize}
                        \item[] Source: "\textit{It was a great evening, until John arrived.}" 
                        \item[] Example of 0: "\textit{It was an evening. The great John arrived.}" 
                        \item[] Example of 1: "\textit{It was a great evening. John arrived.}" 
                        \item[] Example of 2: "\textit{It was a great evening. Until John arrived.}"        
                    \end{itemize}
            \end{itemize}
    
        \item \textbf{Is it grammatical? (scale of 0-2)}          
            \begin{itemize}
                \item Indicate:
                    \begin{itemize}
                        \item 0 if the SPRP hypothesis contains major grammar errors, which impede text comprehension; 
                        \item 1 if it contains minor grammatical errors, which barely impede text comprehension; 
                        \item 2 if there are no grammar errors.                        
                    \end{itemize}
                \item Examples:    
                    \begin{itemize}                    
                         \item[] Source: "\textit{There was a treasure, which was buried on the island.}" 
                         \item[] Example of 0: "\textit{There was a treasure which. He buried on the island.}" 
                         \item[] Example of 1: "\textit{There was a treasure. It buried on the island.}" 
                         \item[] Example of 2: "\textit{There was a treasure. It was buried on the island. }"        
                     \end{itemize}        
            \end{itemize}
    
        \item \textbf{Does it miss any existing fact? (yes/no)}          
            \begin{itemize}
                \item Indicate:
                    \begin{itemize}
                        \item \textit{Yes} if the SPRP hypothesis omits information from the source sentence;
                        \item \textit{No} if the SPRP hypothesis does not omit information from the source sentence. 
                    \end{itemize}
                \item Examples:    
                    \begin{itemize}
                        \item[] Source: "\textit{There was a treasure on a Greek island, which was never discovered.}" 
                        \item[] Example of Yes: "\textit{There was a treasure. It was never discovered.}" 
                        \item[] Example of No: "\textit{There was a treasure on a Greek island. It was never discovered.}" 
                    \end{itemize}
            \end{itemize}
        
        \item \textbf{Does it introduce new facts? (yes/no)}          
            \begin{itemize}
                \item Indicate:
                    \begin{itemize}
                        \item \textit{Yes} if the SPRP hypothesis includes information that was not present in the source sentence;
                        \item \textit{No} if the SPRP hypothesis does not include information that was not present in the source sentence. 
                    \end{itemize}
                \item Examples:    
                    \begin{itemize}
                        \item[] Source: "\textit{There was a treasure, which was never discovered.}" 
                        \item[] Example of Yes: "\textit{There was a treasure on a British island. It was never discovered.}" 
                        \item[] Example of No: "\textit{There was a treasure. It was never discovered.}" 
                    \end{itemize}
            \end{itemize}
    
        \item \textbf{Does it have splits in the wrong place or no split at all? (yes/no)}          
            \begin{itemize}
                \item Indicate:
                    \begin{itemize}
                        \item \textit{Yes} if the SPRP hypothesis includes sentence splits in the wrong places or consists of only one sentence;
                        \item \textit{No} if the SPRP hypothesis does not include sentence splits in the wrong places and contains at least two separate sentences;
                    \end{itemize}
                \item Examples:    
                    \begin{itemize}
                        \item[] Source: "\textit{There was a treasure on a Greek island, which was never discovered and remained there forever.}" 
                        \item[] Examples of Yes: \\
                            \textit{"There was a treasure. On a Greek island it was never discovered. It remained there forever."} \\
                            \textit{"There was a treasure on a Greek island, which was never discovered and remained there forever."}
                            
                        \item[] Examples of No: \\
                            \textit{"There was a treasure on a Greek island. It was never discovered. It remained there forever."} \\
                            \textit{"There was a treasure on a Greek island. It was never discovered and remained there forever."}
                    \end{itemize}
            \end{itemize}

        \item \textbf{Should some of its sentences be further split? (yes/no)}          
            \begin{itemize}
                \item Indicate:
                    \begin{itemize}
                        \item \textit{Yes} if the SPRP hypothesis could be further split and rephrased into more sentences.
                        \item \textit{No} if the SPRP hypothesis could not be further split and rephrased into more sentences.
                    \end{itemize}
                \item Examples:    
                    \begin{itemize}
                        \item[] Source: "\textit{There was a treasure on a Greek island, which was never discovered and remained there forever.}" 
                        \item[] Examples of Yes: \\
                            \textit{"There was a treasure on a Greek island. It was never discovered and remained there forever."} \\
                            \textit{"There was a treasure on a Greek island, which was never discovered. It remained there forever."}
                            
                        \item[] Example of No: \textit{"There was a treasure on a Greek island. It was never discovered. It remained there forever."}
                    \end{itemize}
            \end{itemize}           
            
         \end{itemize}

         \item NOTE: Some of the source sentences contain errors, of the types listed below. These cases should be evaluated as if the source sentence was a correct complex sentence:\footnote{These cases had eluded our initial manual review of the data sampled from Wiki-BM. There were 10 such cases in total in the evaluation, and only 2 for the most problematic first case.}

            \begin{itemize}
                \item The source sentence actually contains two sentences, with missing space around the sentence separator, e.g., "xxx.They xxx".
                \item There is a missing comma, e.g. "They are representatives from Tucson, Arizona who xxx".
                \item There is a missing preposition, e.g. "The plan seemed work".

            \end{itemize}
         
    \end{itemize}

\section{Initial Multilingual Results}
\label{appendix:multilingual}

Our results in this work were based on English, for which different LLM variants and datasets were available. As a first approximation in other languages, in this section we provide results for German, French and Spanish, using a multilingual BLOOM 7.1B model\footnote{https://huggingface.co/bigscience/bloom-7b1} \cite{scao2022bloom} to ensure better coverage than with the English-centric Pythia model. We also include English results with the BLOOM model, for comparison purposes. 

\begin{table*}
\centering
\begin{tabular}{lllrrrrrrr}
\toprule
Language &  Train & Test & BESC & BLEU & SARI & add & keep & del & CPL \\
\midrule
English & DeSSE & DeSSE &  89.05	& 82.00	& 69.56	& 45.51	& 92.42	& 70.74	& 94.05 \\
English & BiSECT & BiSECT & 76.18	& 41.50	& 42.10	& 14.90	& 64.45	& 46.95	& 81.72 \\
\midrule
French & DeSSE & DeSSE &  89.30	& 65.00	& 53.52	& 24.79	& 81.18	& 54.59	& 83.93 \\
French & BiSECT & BiSECT* &  80.61	& 27.20	& 38.45	& 3.88	& 56.55	& 54.92	& 80.21 \\
French & BiSECT & BiSECT &  80.60	& 32.60	& 34.89	& 8.28	& 55.50	& 40.87	& 72.55 \\
\midrule
Spanish & DeSSE & DeSSE &  90.67	& 67.10	& 54.30	& 23.95	& 83.28	& 55.67	& 87.70 \\
Spanish & BiSECT* & BiSECT* & 83.02	& 39.30	& 30.00	& 5.46	& 60.16	& 24.38	& 57.03 \\ 
\midrule
German & DeSSE & DeSSE &  87.62	& 54.80	& 44.75	& 16.47	& 72.85	& 44.93	& 72.22 \\
German & BiSECT* & BiSECT* & 78.19	& 26.80	& 26.02	& 1.67	& 47.92	& 28.47	& 63.13 \\

\bottomrule
\end{tabular}
\caption{Results in English, German, French and Spanish with the Bloom-7B1 model. Train and test datasets with missing diacritics are indicated with an asterisk. BertScore results are only normalised for English.}
\label{table:mutlingual-results}
\end{table*}

Since the DeSSE datasets were only available in English, we used an NLLB MT model\footnote{https://huggingface.co/facebook/nllb-200-3.3B}  \cite{costa2022no} to translate the corpora into each language. Although this approximation can introduce translation errors in the process, it was the only available alternative for the DeSSE scenario. 

The BiSECT corpora were available for all three of the aforementioned languages, but suffered from key issues: to the exception of the French training corpus, all datasets were missing diacritics, which are critical in these languages. To partially cope with this issue, for French we created an additional test set by randomly extracting 2,000 pairs from the original training dataset; for the other languages, we report results using the faulty datasets, as there was no alternative in these cases. Train and dev size were the same as for English, whereas the test sets contained 1036, 2018 and 735 pairs for French, Spanish and German, respectively.


The results for these experiments are summarised in Table~\ref{table:mutlingual-results}. The English results obtained with the BLOOM model were similar to those obtained with the Pythia model (see Table~\ref{table:comp-results}), indicating that the overall SPRP ability does not seem impacted by variance between fine-tuned models of similar overall architecture and parameter size.

On the machine-translated DeSSE datasets, the results were high overall across metrics for Spanish, French, German, in descending accuracy order, although below those obtained in English, presumably as a result of the comparatively smaller volumes of LLM training data in languages other than English. On BiSECT, of note are the results obtained with models trained and tested on datasets that lacked diacritics. These models still generated output that included correct word forms with diacritics, the underlying model having been trained on correct data. The scores obtained against test sets without diacritics were thus negatively affected, and all models but the French one were trained on faulty data, an additional artificial constraint. The results in French on the test set with diacritics are below English but within range overall. They are also below the results on the faulty test set in terms of SARI and compliance, but these differences may be due to the non-faulty test set being twice the size of the original one.

Overall, these results tend to indicate strong potential of multilingual LLMs for the SPRP task, factoring the lack of, or specific issues with, existing datasets in languages other than English.

\end{document}